\documentclass[letterpaper, 10pt, conference]{ieeeconf}  

\IEEEoverridecommandlockouts 

\usepackage{amsmath,amsfonts,amssymb,mathtools} 
\usepackage{bbm,bm,nicefrac} 
\usepackage{graphicx,float} 
	\graphicspath{{./images/}}
\usepackage[table,xcdraw]{xcolor}
\usepackage{algorithm,algpseudocode} 
\let\labelindent\relax \usepackage[inline]{enumitem}
\usepackage{cite}
\usepackage{cases}
\usepackage{bm}
\usepackage{soul,color}

\newtheorem{remark}{Remark}

\newcommand{\T}{\top} 

\algnewcommand{\algorithmicand}{\textbf{ and }}
\algnewcommand{\algorithmicor}{\textbf{ or }}
\algnewcommand{\OR}{\algorithmicor}
\algnewcommand{\AND}{\algorithmicand}
\algnewcommand{\var}{\texttt}

\title{\LARGE \bf RCMS: Risk-Aware Crash Mitigation System for Autonomous Vehicles}

\author{Faizan M. Tariq$^{1}$, David Isele$^{2}$, John S. Baras$^{1}$ and Sangjae Bae$^{2}$
\thanks{$^{1}$University of Maryland, College Park, MD, USA. Email: \tt\small\{mftariq,baras\}@umd.edu.}
\thanks{$^{2}$Honda Research Institute, San Jose, CA, USA. Email: \tt\small\{disele,sbae\}@honda-ri.com.}
\thanks{Research supported by Honda Research Institute, USA.}
}

\begin{document}

\maketitle
\begin{abstract}
We propose a risk-aware crash mitigation system (RCMS), to augment any existing motion planner (MP), that enables an autonomous vehicle to perform evasive maneuvers in high-risk situations and minimize the severity of collision if a crash is inevitable. In order to facilitate a smooth transition between RCMS and MP, we develop a novel activation mechanism that combines instantaneous as well as predictive collision risk evaluation strategies in a unified hysteresis-band approach. For trajectory planning, we deploy a modular receding horizon optimization-based approach that minimizes a smooth situational risk profile, while adhering to the physical road limits as well as vehicular actuator limits. We demonstrate the performance of our approach in a simulation environment.
\end{abstract}

\section{Introduction}
Many modern vehicles already contain collision warning and braking systems that help to reduce the number and severity of rear-end collisions \cite{emergencyBraking, fuzzyRear}. However, in terms of control, these systems are limited strictly to braking behaviors. This work looks at developing a more complete steering and acceleration control system capable of reducing the number of collisions in a wider class of situations. This system could potentially be used, like collision warning and braking systems, as an advanced driver assistance system (ADAS) or it could be coupled with full autonomous driving (AD) software system as a fail-safe protection.

\begin{figure} [ht]
\centering
\includegraphics[trim=0 0 0 0, clip,width=.4\textwidth]{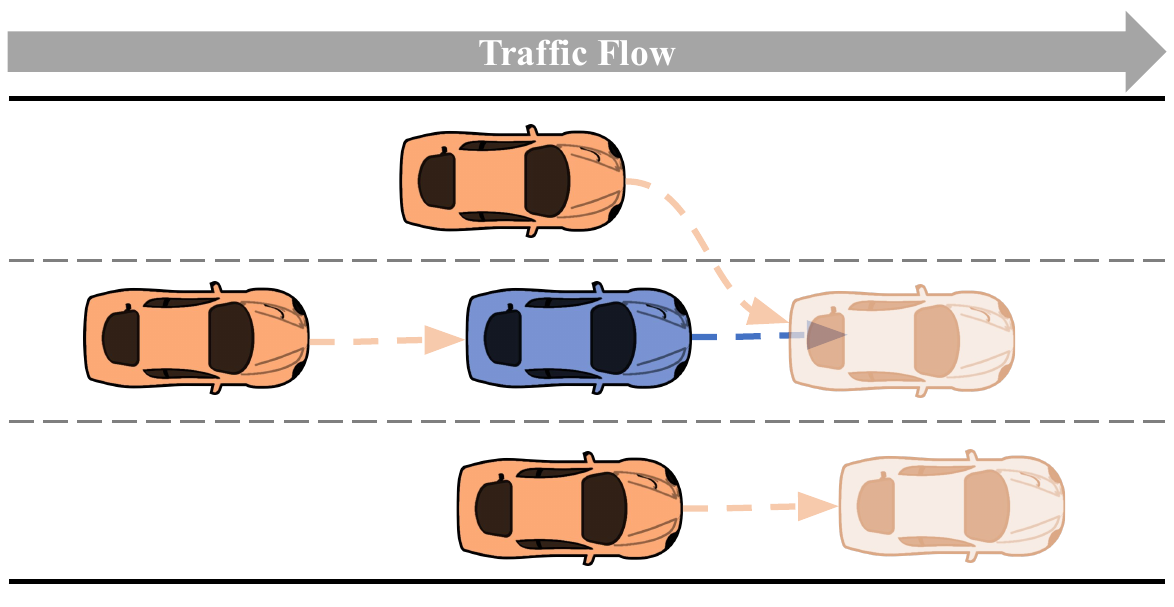}
\caption{\textbf{Motivational Example}. The ego vehicle (in blue) is cruising on a highway, with a vehicle following behind, and two vehicles traveling on either side in adjacent lanes. If one of the vehicles in the adjacent lanes suddenly swerves towards it, the ego vehicle is put in a high-risk situation where a collision may be unavoidable. The ego vehicle will have to make an evasive maneuver that ideally avoids a crash but if it's inevitable, then choose an action that minimizes the severity of collision.}
\label{fig:overview}
\vspace{-15pt}
\end{figure}

To motivate our work, consider the scenario presented in Fig. \ref{fig:overview}, where an ego vehicle has vehicles traveling on its side as well as behind. If one of the vehicles on the adjacent lanes swerves into the ego vehicle's lane (possibly due to a blind spot), the ego vehicle has to perform a drastic maneuver to avoid a collision, or minimize its severity, if it is unavoidable. It cannot simply use emergency braking \cite{emergencyBraking} due to the trailing vehicle, and it cannot swerve into the other lane due to the presence of the other vehicle. Therefore, it has to finesse its way around the vehicles, depending on the space availability and actuation limits, to take the least risky action. To handle such intricate scenarios, we develop a risk-aware crash mitigation system (RCMS) in this work.

The design of RMCS involves two components: \begin{enumerate*}[label=(\roman*)] \item a novel activation mechanism and, \item a trajectory generation method \end{enumerate*}. The activation mechanism considers both the instantaneous as well as predictive collision risk evaluation strategies under a hysteresis band to trigger the trajectory generation method. The trajectory generation method performs situational risk analysis through smooth functional evaluation and optimizes it in a receding horizon optimization-based framework. Due to its modular nature, RCMS has the capability to augment any existing motion planning framework while incorporating modern prediction algorithms.

While crash mitigation in an AD setting depends also on timely detection \cite{huijser2009animal} and prediction \cite{muller2018machine}, we focus on the planning and decision-making aspect of the problem. In that regard, most existing literature focuses on risk-based techniques.
Lee and Kum \cite{samplingPOM} evaluate situational risk through a predictive occupancy map (POM) and use a sampling-based technique for trajectory generation. They employ a simplistic threshold-based activation mechanism and a fixed time-based deactivation. The sampling-based approach restricts the solution search space to the set of predetermined samples which may be very limiting in high-risk scenarios, where the difference between a trajectory that avoids a collision to the one that does not may be minimal. Moreover, the fixed time-based deactivation runs the risk of deactivating the system before getting the vehicle to a safe state or leaving it in an even worse situation during the transition.

Wang et al. \cite{wang2019crash} present a real-time Model Predictive Control (MPC) algorithm that uses a potential crash severity index (CSI) to select the least dangerous action. However, in an effort to improve the computational complexity, they linearize the dynamical model and subsequently convexify the optimization problem, potentially adversely affecting the feasibility of the control actions, which may prove detrimental in a collision avoidance situation. Moreover, their work does not consider when to activate the system, limiting its usefulness for ADAS applications.

Shang et al. \cite{shang2023emergency} combine artificial potential fields with MPC and verify that against a Hamilton Jacobi reachability (HJ) based approach. They use a \emph{non-smooth} energy-based cost function which may adversely affect the computational complexity, as no timing statistics for the algorithm are provided, while the HJ-based approach notably has high computational complexity, so a simpler unicycle dynamical model is used. Moreover, their relatively simple rule-based activation mechanism is prone to running into issues pertaining to ineffective triggering, as discussed in Section \ref{sec:activation}.

Qin et al. \cite{integratedCrashAvoidanceMitigation} integrate a high-fidelity model with tire slip forces while considering MAIS (Maximum Abbreviated Injury Severity) 3+ probability as CSI while using time-to-collision (TTC) to switch between three modes of operation: path following, crash avoidance, and crash mitigation. Although promising, such an approach raises concerns regarding its real-time applicability, as no timing statistics were provided in the paper. Furthermore, as mentioned previously, using a simplistic condition-based triggering mechanism is not adequate for effective switching between the crash mitigation system and the regular motion planner.

In this work, we address the aforementioned drawbacks of existing methods by designing RCMS, composed of an activation mechanism and a modular trajectory generation component. The activation mechanism combines instantaneous as well as predictive collision risk evaluation in a hysteresis band to facilitate a smooth transition between RCMS and MP, which is important since the goals of the two systems are fundamentally different, as discussed in Section \ref{sec:activation}. The trajectory generation component minimizes the situational risk, evaluated through a smooth function while considering actuation, dynamical, and road limits, as detailed in Section \ref{sec:trajectoryGeneration}. We verify the performance of our approach while providing timing statistics to ascertain its real-time applicability in the simulation of high-risk collision-prone scenarios in Section \ref{sec:results}.

\section{System Overview}
In this section, we elaborate upon the algorithmic pipeline as well as the road, observation, and vehicle models.

\subsection{Algorithmic Pipeline}
\label{sec:pipeline}

\begin{figure} [htb]
\centering
\includegraphics[trim=0 0 0 0, clip,width=0.8\columnwidth]{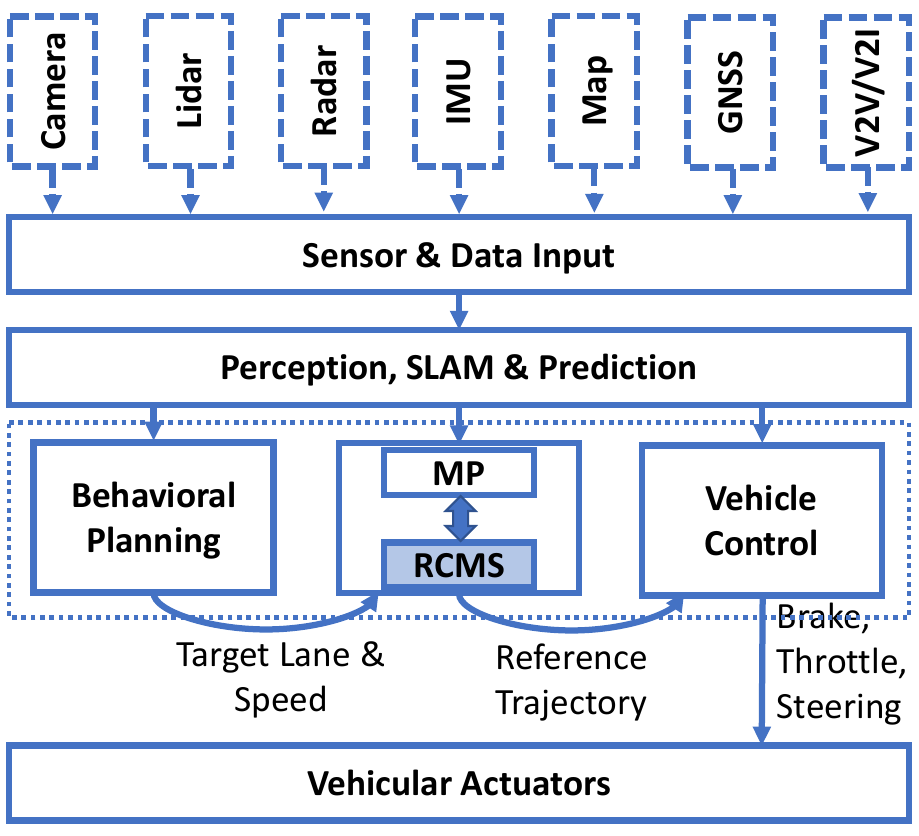}
\caption{\textbf{Algorithmic Pipeline}. The Perception and Simultaneous Localization and Mapping (SLAM) modules process the raw sensory input to locate the AV within the environment. The navigation stack (outlined by the dotted rectangle), composed of behavioral planning, motion planning, and vehicle control modules, uses this information to control the AV through actuation commands (brake, throttle, and steering).
}
\vspace{-15pt}
\label{fig:pipeline}
\end{figure}

Fig.~\ref{fig:pipeline} provides a visual representation of the data flow among various algorithmic modules onboard an autonomous vehicle (AV). The navigation stack, enclosed by the dotted rectangle, consists of several modules whose nomenclature follows the conventions established in \cite{surveyPlanning}. This modular architecture facilitates the seamless integration of external modules while maintaining the overall system integrity. Additional insights into this framework can be found in our previous work \cite{slas}. The primary objective of this research is to augment any existing motion planning module with a crash mitigation system, as depicted in Fig.~\ref{fig:pipeline}.

\subsection*{Notation} 
Throughout the manuscript, we denote the set of integers by $\mathbb{Z}$ and the set of real numbers by $\mathbb{R}$. For some $a,c \in \mathbb{Z}$ and $a<c$, we write $\mathbb{Z}_{[a,c]} = \{b \in \mathbb{Z} \mid a \leq b \leq c \}$. For some $e,g \in \mathbb{R}$ and $e<g$, we write $\mathbb{R}_{[e,g]} = \{f \in \mathbb{R} \mid e \leq f \leq g \}$. We reserve the variable $k \in \mathbb{R}$ to represent the current time instant such that any variable defined as $\zeta(k)$ represents the evaluation of $\zeta$ at the current instant.

\subsection{Road Model} \label{sec:roadModel}
Even though the proposed methodology can be applied to any general setting, we limit the scope of this work to a multi-lane highway setting, as depicted in Fig. \ref{fig:overview}.
In this setting, we define spatial coordinates $(x,y)$ with respect to a global Cartesian coordinate system.
We assume that the road limits $\mathbb{B}(k) \subset \mathbb{R}^2$ in this coordinate system as well as the speed limit $V(k)$ can be measured.

\subsection{Observation Model}
We limit the ego vehicle's visibility range to its field of view, denoted by $\mathbb{F}(k)$. With the set of vehicles in the environment represented by $\mathbb{E}(k) \subset \mathbb{Z}_{>0}$, the set of vehicles visible to the ego vehicle is given by:
\begin{equation}
\mathbb{O}(k) = \left\{i \in \mathbb{E}(k) ~\big|~ p_i(k)  \in \mathbb{F}(k) \right\}
\end{equation}
where $p_i(k) = \begin{bmatrix} x_i(k) & y_i(k) \end{bmatrix}^\T$ denotes the position of vehicle $i$'s center of mass in the global Cartesian coordinate frame, with subscript $_0$ reserved for the ego vehicle.

\begin{remark}
    Active perception through the receding horizon approach (Section \ref{sec:trajectoryGeneration}) enables the ego vehicle to handle challenges pertaining to partial information, resulting from occlusion, sensory limitations, etc., as demonstrated in \cite{overtakingBidirectional}. Moreover, various streams of input data, e.g. V2X communication methods, can provide additional information to overcome such challenges, as shown in \cite{cooperativeOvertaking}.
\end{remark}

\subsection{Vehicle Model}
\label{sec:vehicleModel}
We utilize the nonlinear kinematic bicycle model which provides a good balance between efficiency and accuracy \cite{bicycleModel}. Thus, the states, $X(k)$, and control inputs, $U(k)$, of the ego vehicle at time instant $k$ are defined as:
\begin{align}
    X(k) &=
    \begin{bmatrix} 
    x_0(k) & y_0(k) & \theta_0(k) & v_0(k)
    \end{bmatrix}^\T \in \mathbb{X}(k), \\
    U(k) &= \begin{bmatrix}
    a(k) & \delta(k)
    \end{bmatrix}^\T \in \mathbb{U}(k),
\end{align}
where $x_0(k)$, $y_0(k)$, $\theta_0(k)$, and $v_0(k)$ respectively denote the x-coordinate ($m$), y-coordinate ($m$), yaw angle with respect to the x-axis ($rad$), and speed ($m/s$), whereas $a(k)$ and $\delta(k)$ respectively denote acceleration ($m/s^2$), and steering angle ($rad$). The sets $\mathbb{X}(k) = \mathbb{B}(k) \times \mathbb{R}_{[0, 2\pi)} \times \mathbb{R}_{[0,2V(k)]}$ and $\mathbb{U}(k) = \mathbb{R}^2_{[U_{\text{min}}, U_{\text{max}}]}$ respectively denote the feasible states and actuation limits. Then, the system dynamics read:
\begin{equation}    \label{eqn:dynamics}
\begin{aligned}
    x_0(k+1) &= x_0(k) + T_s \cdot (v_0(k) \cos (\theta_0(k)),\\
    y_0(k+1) &= y_0(k) + T_s \cdot (v_0(k) \sin (\theta_0(k)),\\
    \theta_0(k+1) &= \theta_0(k) + T_s \cdot \left(\frac{v_0(k)}{L} \tan(\delta(k))\right),\\
    v_0(k+1) &= v_0(k) + T_s \cdot a(k),
\end{aligned}
\end{equation}
where $T_s$ corresponds to the sampling time. Further details regarding this model can be found in \cite{bicycleModel}.

\begin{remark}
    The set of feasible speeds is expanded to twice the speed limit as it may be necessary to exceed the speed limit in order to avoid a crash and maintain safety.
\end{remark}

\begin{remark}
    In our dynamical model, we do not incorporate the tire model or friction forces \cite{tireModel}. Although these factors may play a significant role in extreme situations, we have to be mindful of the efficiency-accuracy tradeoff. Since our focus here is on the motion planning layer, rather than the control layer (refer to Fig. \ref{fig:pipeline}), where challenges arise from the inclusion of collision avoidance constraints (Section \ref{sec:trajectoryGeneration}) and a longer planning horizon, incorporating higher fidelity models comes at a cost of degraded computational efficiency.
\end{remark}

\section{Approach}
\label{sec:approach}
This section describes the activation mechanism responsible for triggering the RCMS and the receding horizon optimization-based trajectory generation module, together with its various components, that outputs a reference trajectory to the low-level modules.

\subsection{Activation Mechanism} \label{sec:activation}
An essential component of RCMS is its activation mechanism which decides when and how to activate it. This is because the objective of RCMS is fundamentally different from that of MP which operates under normal (low-risk) operating conditions. Unlike MP, RCMS places minimal emphasis on auxiliary metrics such as waypoint following, travel time minimization, passenger comfort maximization, etc., and focuses solely on the fundamental need to ensure the safety of the ego vehicle. 



Being cognizant of the underlying differences in the objectives of the two modules and the potential undesirable consequences of improper switching, we propose a novel activation mechanism that facilitates a smooth transition between the two systems. In the existing literature, we notice the use of either instantaneous or predictive risk evaluation strategies to determine the necessity to activate the crash mitigation system. In terms of the instantaneous and predictive risk evaluation methods respectively, Gaussian overlap \cite{riskGaussian} and time-to-collision (TTC) \cite{fuzzyRear, integratedCrashAvoidanceMitigation, threatAssessment} appear to be the preferred choices of the research community due to their simplicity and efficiency. However, these methods are usually deployed standalone in a `bang-bang' fashion which runs the risk of constant switching between the two systems, adversely affecting passenger comfort as a result. Moreover, having them deployed independently also runs the risk of underestimating the risk in certain situations due to their underlying formulations. To elaborate, consider the following two scenarios. In the first case, the ego vehicle is traveling behind a human-driven vehicle (HDV) with negligible headway and 0 relative velocity, leading to $\infty$-TTC. If the leading vehicle decides to brake suddenly, it will lead to a crash so this is a high-risk situation that is not captured by the TTC metric but is captured by the Gaussian overlap metric due to the close proximity of the two vehicles. On the other hand, consider the previous scenario but with considerable headway between the two vehicles. In this case, if the HDV decides to brake, the Gaussian overlap will not have a high enough value until the two vehicles get into close proximity of each other, but by that time, it may be too late to take any evasive actions, so TTC, with its predictive nature, has a better chance of anticipating the developing high-risk situation. To counter these drawbacks and ensure smooth operation, we combine instantaneous and predictive risk evaluation methodologies in a unified hysteresis-based activation mechanism.

As for instantaneous collision risk evaluation, motivated by \cite{riskGaussian}, we model the risk associated with vehicle $i$ as a bivariate Gaussian distribution $\tilde p_i(k) \sim \mathcal{N}(p_i(k),\Sigma_i(k))$. Here, the mean corresponds to the position of the vehicle while the variance matrix is determined by the length, width, and orientation of the vehicle as follows:
\begin{equation}
\label{eqn:covariance}
    \Sigma_i(k) = R_{\theta_i(k)} \begin{bmatrix}
        \beta_l L_i & 0 \\
        0 & \beta_w W_i
    \end{bmatrix} R_{\theta_i(k)}^\T,
\end{equation}
where $L_i$ and $W_i$ define the length and width of vehicle $i$ with $\beta_l$ and $\beta_w$ corresponding to the respective scaling factors while $R_{\theta_i(k)}$ represents the 2D rotation matrix with the rotation angle $\theta_i(k)$.

We then evaluate the ego vehicle's overall collision risk $\kappa(k)$ as the maximum of its pairwise collision risk, $\kappa_i(k)$, with vehicle $i$ where $\kappa_i(k)$ is given as the \emph{product sum} of their distributions \cite{riskGaussian}. This is analogous to likelihood since when two distributions are completely overlapped, the likelihood is highest and when the distributions are not overlapped, the likelihood is very low. We evaluate $\kappa(k)$ analytically as follows:
\begin{align}
    \kappa(k) &= \max_{i \in \mathbb{O}(k)} \kappa_i(k), \label{eqn:collisionRisk}\\
    \label{eqn:pairwiseCollisionRisk}
    \kappa_i(k) &= \eta_i(k) \cdot e^{\frac{1}{2}\left(\omega_i^\top\Omega_i^{-1}\omega_i-p_0^\top\Sigma_0^{-1}p_0-p_i^\top\Sigma_i^{-1}p_i\right)},
\end{align}
where $\Omega_i=\left( \Sigma_0^{-1} + \Sigma_i^{-1}\right)$, $\omega_i=\Omega_i\left(\Sigma_0^{-1}p_0+\Sigma_i^{-1}p_i\right)$, and $\eta_i(k)$ is a normalizing/scaling factor.

\begin{remark}
    For the sake of brevity, we do not show the dependence of variables in (\ref{eqn:pairwiseCollisionRisk}) on time instant $k$. 
\end{remark}

\begin{remark}
    To keep the formulation generalized, we let $\eta_i(k)$ be vehicle $i$ and time $k$ dependent which gives us the flexibility to modify the risk based on a vehicle's class, e.g. truck, emergency, bicycle, etc., and its behavior over time. 
\end{remark}

\begin{figure}
\centering
\includegraphics[trim=0 0 0 0, clip,width=.425\columnwidth]{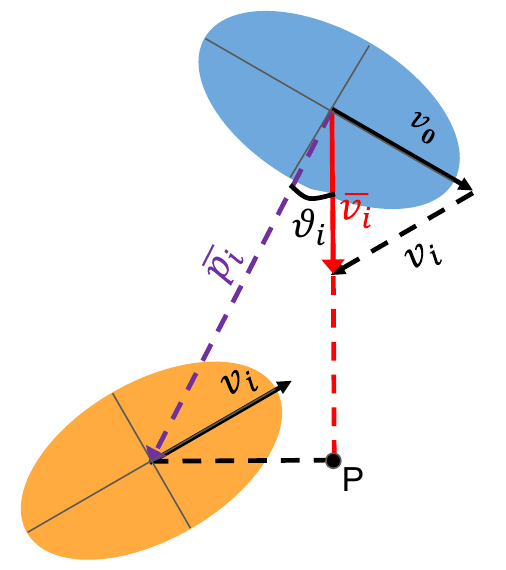}
\caption{\textbf{Time-to-Closest-Encounter (TTCE) Calculation}. With the ego vehicle shown in blue and the vehicle $i$ shown in orange, the TTCE is calculated based on the relative velocity ($\bar v_i$) as well as the relative displacement ($\bar p_i$) between the two vehicles.}
\vspace{-10pt}
\label{fig:ttc}
\end{figure}

Regarding the predictive collision risk metric, TTC is typically evaluated based only on longitudinal displacement between vehicles \cite{fuzzyRear}. However, for the crash mitigation system in a highway setting, lateral motion plays as much of a role, if not more, as longitudinal motion. Due to the direction of motion in the longitudinal direction, the motion planner already tends to maintain a desired safety margin in that direction, but the lane width typically restricts the safety margin in the lateral direction. For instance, it is easier to anticipate and react to the behavior of a vehicle traveling further ahead, even if it is acting erratically, but it's harder to react to a vehicle traveling in the adjacent lane that suddenly decides to swerve in the direction of the ego vehicle. Therefore, we formulate our predictive collision risk evaluation metric $\tau (k)$, based on the pairwise Time-to-Closest-Encounter (TTCE) $\tau_i (k)$, in terms of the relative velocity between the ego vehicle and vehicle $i$ as follows:
\begin{align}
    \tau(k) &= \max_{i \in \mathbb{O}(k)} \frac{1}{\tau_i(k)} \label{eqn:ttce}\\
    \tau_i(k) &= \frac{\|\bar p_i(k)\| \cos(\vartheta_i(k))}{\| \bar v_i(k)\|} = \frac{\bar p_i(k) \cdot \bar v_i(k)}{\|\bar v_i(k)\|^2} \label{eqn:ttce_i}
\end{align}
where $\bar p_i(k) = p_i(k) - p_0(k)$ and $\bar v_i(k)$ correspond respectively to the relative displacement and velocity between the ego vehicle and vehicle $i$, as shown in Fig. \ref{fig:ttc}, while the second equality in (\ref{eqn:ttce_i}) follows from the fact that:
\begin{equation*}
    \displaystyle \bar p_i(k) \cdot \bar v_i(k) = \|\bar p_i(k)\| \|\bar v_i(k)\| \cos(\vartheta_i(k)).
\end{equation*}

\begin{figure} [!ht]
\centering
\includegraphics[trim=0 0 0 0, clip,width=.325\textwidth]{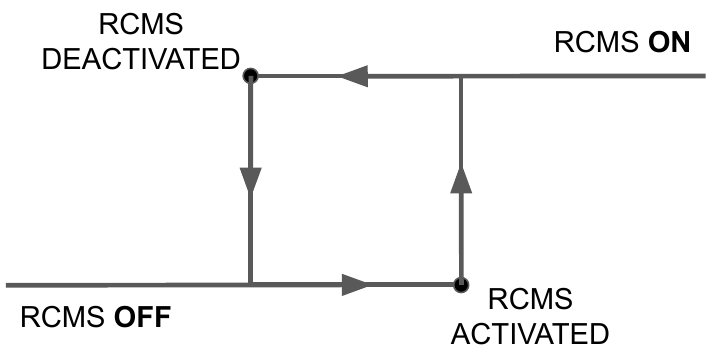}
\caption{\textbf{Hysteresis Band for RCMS Activation}. Different activation and deactivation thresholds enable a much smoother transition between MP and RCMS compared to a `bang-bang' approach.}
\label{fig:hysteresis}
\vspace{-10pt}
\end{figure}

We combine the instantaneous (\ref{eqn:collisionRisk}) and predictive collision risk evaluation metrics with a hysteresis band, depicted in Fig. \ref{fig:hysteresis}, to come up with an overall activation mechanism, outlined in Algorithm \ref{alg:activation}. With the hysteresis band, higher activation thresholds ($\kappa_a$ and $\tau_a$) prevent unnecessary activation of the crash mitigation system while lower deactivation thresholds ($\kappa_d$ and $\tau_d$) allow for the ego vehicle to get to a much safer state before handing the control back over to the motion planner. Moreover, the two conditions in line~\ref{alg_ln:predictiveCondition} of Algorithm~\ref{alg:activation} ensure that the predictive collision risk is evaluated only when necessary. Specifically, $\bar v_i(k) \cdot \bar p_i(k) < 0$ ensures that the ego vehicle and vehicle $i$ are moving towards each other while $ \| \bar p_i(k) \| \sin(\vartheta_i(k)) < L_0 + L_i + \epsilon$ ensures that at the point of closest encounter (`P' in Fig.~\ref{fig:ttc}), the vehicles are in close proximity to each other where $\epsilon$ governs how close of a proximity to consider. In the 2D Cartesian coordinate system, we can evaluate $\| \bar p_i(k) \| \sin(\vartheta_i(k))$ using the following relation:
\begin{align*}
    \bar p_i(k) \times \bar v_i(k) &= \|\bar p_i(k)\| \|\bar v_i(k)\| \sin(\vartheta_i(k)) \\
    \|\bar p_i(k)\| \sin(\vartheta_i(k)) &= \frac{\bar p_i(k) \times \bar v_i(k)}{\|\bar v_i(k)\|}.
\end{align*}

\begin{algorithm}
\caption{RCMS Activation Mechanism}\label{alg:activation}
\begin{algorithmic}[1]
\State $\kappa(k) \gets 0$
\State $\tau(k) \gets \infty$
\While {system running}
    \For{$i \in \mathbb{O}(k)$}
        \State $\kappa_i(k) \gets (\ref{eqn:pairwiseCollisionRisk})$
        \If {$\bar v_i(k) \cdot \bar p_i(k) < 0$ \AND \newline \hspace*{2.5em}
        $ \| \bar p_i(k) \| \sin(\vartheta_i(k)) < L_0 + L_i + \epsilon$} \label{alg_ln:predictiveCondition} 
            \State $\tau_i(k) \gets (\ref{eqn:ttce})$
        \Else{}
            \State $\tau_i(k) \gets \infty$
        \EndIf
    \EndFor
    \State $\kappa(k) \gets (\ref{eqn:collisionRisk})$
    \State $\tau(k) \gets (\ref{eqn:ttce})$
    \If {$\kappa(k) > \kappa_a \OR \tau(k) > \tau_a$}
        \State Activate RCMS
    \ElsIf{$\kappa(k) < \kappa_d \AND \tau(k) < \tau_d$}
        \State Deactivate RCMS
\EndIf
\EndWhile
\end{algorithmic}
\end{algorithm}

\subsection{Trajectory Generation} \label{sec:trajectoryGeneration}
Once the crash mitigation system has been activated, its job, just like that of the motion planner, is to provide a reference trajectory comprised of future waypoints (position as well as speed profile), for the low-level modules (see Fig. \ref{fig:pipeline}) to track and follow. For trajectory generation, we utilize a receding horizon optimization \cite{mpcDriving} based formalism that allows us to minimize the situational risk profile while accounting for vehicular dynamics, actuator limits, and road boundaries, ensuring the feasibility of output trajectory.

\subsection*{Situational Risk Model}
To quantify the spatially distributed situational risk perceived by the ego vehicle at time instant $k$, we formulate the instantaneous risk function $\rho(k)$ as an aggregate of individual agent-specific risk functions $\rho_i(k)$, for each $i \in \mathbb{O}(k)$, as well as road boundary risk $\rho_r(k)$. We choose parametric continuous smooth functions to model $\rho(k)$. In particular, taking inspiration from a skewed Gaussian distribution \cite{skewNormal}, we formulate a skewed hyperbolic quadratic function to represent $\rho_i(k)$, and we model $\rho_r(k)$ with a univariate Gaussian function. Before expanding on the functional form of $\rho_i(k)$, we first shed some light on the density function $\phi_s(p_s)$ of a skewed Gaussian distribution, outlined below:
\begin{equation}
    \phi_s(p_s) = 2 \phi(p_s; \mu_s, \Sigma_s) \Phi(q_s^\T p_s),
\end{equation}
where $p_s \in \mathbb{R}^2$ and $q_s \in \mathbb{R}^2$ correspond respectively to the position and direction-oriented skew parameter in the Cartesian coordinate frame while $\phi$ and $\Phi$ represent the density and distribution functions of a bivariate Gaussian distribution parametrized by mean $\mu_s$ and covariance $\Sigma_s$.

To obtain a situational risk profile with its spatial density similar to that of a Gaussian mixture while ensuring that each agent has a barrier around that prevents the ego vehicle from `going through' it, we opt to use a reciprocal quadratic function $\psi_i$ instead of $\phi$. Moreover, to orient and scale the symmetric risk distribution in the direction of an agent's motion, we use a simpler Sigmoid function $\sigma_i$ instead of $\Phi$ which requires the evaluation of the error function. Then, having $\bar p(k) = \{\bar p_i(k) \mid i \in \mathbb{O}(k) \}$, $\rho(k)$ is given by:
\begin{align}
    \rho(k; \bar p(k)) &= \sum_{i \in \mathbb{O}(k)} \rho_i(k; \bar p_i(k)) \label{eqn:riskSituational} + \rho_r(k) \\
    \rho_i(k; \bar p_i(k)) &= \psi_i(k; \bar p_i(k)) \sigma_i(k; \bar p_i(k)) \\
    \psi_i(k; \bar p_i(k)) &= \frac{\tilde \eta_i(k)}{\alpha_g + {\bar p_i^\T(k)} \widetilde \Sigma_i ^{-1}(k) \bar p_i(k)} \\
    \sigma_i(k; \bar p_i(k)) &= \frac{1}{1+\exp(-\alpha_s \bar p_i^\T(k) v_i(k))} \\
    \rho_r(k) &= \gamma_r \exp \left( - \alpha_r {\bar p_r^\T(k) \Gamma(k) \bar p_r(k)} \right)
\end{align}
where $\alpha_g$ controls the relative gradient of the agent distribution; $\alpha_s$ controls skewness of the agent distribution; $\tilde \eta_i(k)$ serves a similar normalization purpose to $\eta_i(k)$ in (\ref{eqn:pairwiseCollisionRisk}); $\widetilde \Sigma_i(k)$ is defined analogously to $\Sigma_i(k)$ in (\ref{eqn:covariance}) but the distinction is made since the scaling and rotation parameters need not be the same; $\gamma_r$ controls the scaling for road distribution; $\alpha_r$ controls the gradient for road distribution; $\bar p_r(k) = p_r(k) - p_0(k)$, with $p_r(k) \in \mathbb{B}(k)$, denotes the relative coordinates of road boundary; and $\Gamma (k)$, with the help of Frenet coordinates system, ensures that the road risk is effective only in the lateral direction.



\subsection*{Prediction Model}

Given the extensive literature on motion prediction methods for autonomous vehicles \cite{surveyPrediction}, there exists a range of approaches that can be integrated into the RCMS framework. However, the lack of data for high-risk near-miss or collision scenarios raises doubts about the applicability of many state-of-the-art methods. To avoid making assumptions about the nature of neighboring agents as it may lead to degraded overall behavior, we opt to use a constant acceleration prediction model for RCMS. Making assumptions about the behavior of agents, such as attributing aggressive (adversarial) or defensive (cooperative) intent to them, increases the risk of misjudging the true nature of the agent's behavior in such stressful scenarios. This can lead to undesirable outcomes, such as focusing on minimizing the severity of a crash but ultimately causing the crash when it was actually avoidable or opting for a collision avoidance strategy but eventually increasing crash severity when it was indeed unavoidable. In practice, most high-risk situations result from human drivers' negligence \cite{humanError, crashStats} which justifies the use of our simple behavior-agnostic prediction model formalized below:
\begin{align}
    x_i^k(j+1) &= x_i^k(j) + T_s v_i^k(j) \cos(\theta_i(k))\\
    y_i^k(j+1) &= y_i^k(j) + T_s v_i^k(j) \sin(\theta_i(k)) \\
    v_i^k(j+1) &= v_i^k(j) + T_s a_i(k)
\end{align}
where $\theta_i(k)$ and $a_i(k)$ are vehicle $i$'s estimated heading and acceleration values at time instant $k$ while $x_i^k(j)$, $y_i^k(j)$ and $v_i^k(j)$ are vehicle $i$'s predicted x-coordinate, y-coordinate and speed values at a future time step $j$ w.r.t. time instant $k$, with $x_i^k(0) = x_i(k)$, $y_i^k(0) = y_i(k)$ and $v_i^k(0) = v_i(k)$. Then, the relative predicted position is denoted by $\bar p_i^k(j) = \begin{bmatrix}
    x_i^k(j) & y_i^k(j)
\end{bmatrix}^\T - p_0(k)$ and $\bar p^k(j) = \{\bar p_i^k(j) \mid i \in \mathbb{O}(k) \}$.

\begin{remark}
    The modular nature of RCMS allows for the incorporation of modern prediction algorithms in case better models become available for the task at hand.
\end{remark}

\begin{figure*} [ht]
\centering
\includegraphics[trim=0 0 0 0, clip,width=0.95\textwidth]{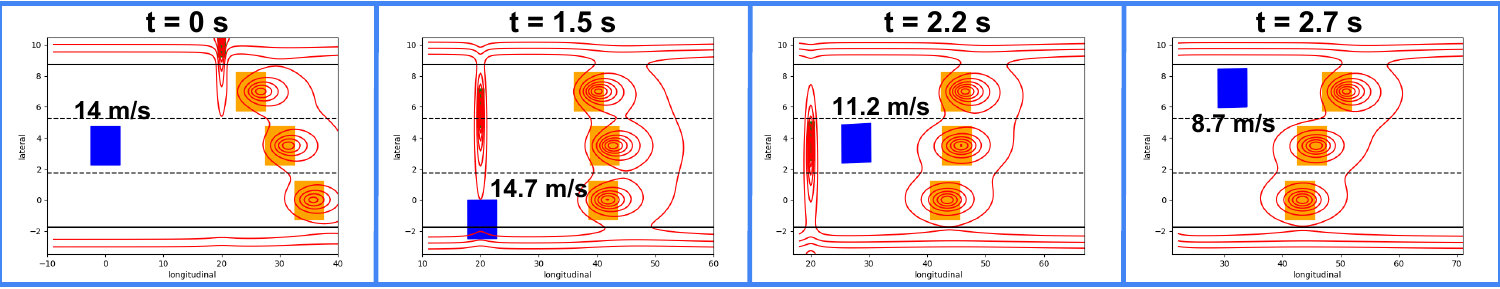}
\caption{\textbf{Scenario I.} With the on-road agents depicted in orange, and the situational risk contour plot depicted in red, the ego vehicle (in blue) notices a moving object (animal, pedestrian etc.) approaching laterally from the end of the road, leading to a high collision risk, which activates the RCMS at $t=0s$. The ego vehicle then swerves right to avoid a collision with the object when suddenly two of the vehicles traveling ahead in the right and center lanes successively stop abruptly in the middle of the highway at $t=1.5s$, maintaining the high collision risk. Then, the ego vehicle swerves smoothly to the left-most lane to place the ego vehicle in a relatively safe state before handing the control over to the MP at $t=2.7s$.}
\label{fig:scenario_animal}
\end{figure*}
\begin{figure*} [ht]
\centering
\includegraphics[trim=0 0 0 0, clip,width=0.95\textwidth]{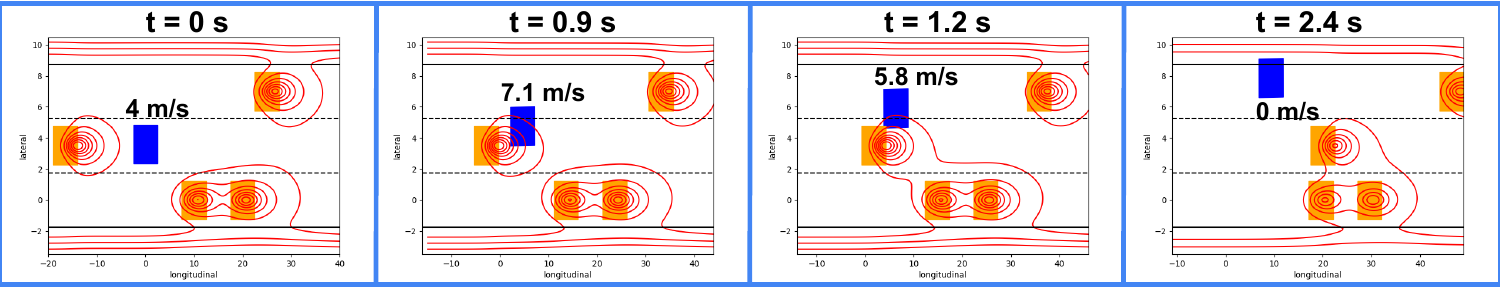}
\caption{\textbf{Scenario II.} With the on-road agents depicted in orange, and the situational risk contour plot depicted in red, the ego vehicle (in blue) traveling slowly in the center lane notices a fast-moving vehicle approaching rapidly from behind, leading to a high collision risk, which activates the RCMS at $t=0s$. With two other vehicles traveling close up ahead towards its right, the ego vehicle accelerates and swerves left simultaneously to minimize the severity of the collision, leading to a near-miss situation at $t=1.2s$. Due to the drastic maneuver, the ego vehicle has to make use of the shoulder to stabilize before handing the control back to the MP at $t=2.4s$.}
\label{fig:scenario_vehicle}
\vspace{-5pt}
\end{figure*}

\subsection*{Objective Function}
The objective function is formulated to minimize the accumulative predictive situational risk, defined in (\ref{eqn:riskSituational}), over the planning horizon $H$ while placing relatively low emphasis on control actions regulation as the priority is to ensure safety.
%
\begin{equation}
    J(k) = \sum_{j=1}^H \rho^k(j; \bar p^k(j)) + {U^k}^\T(j) R(k) U^k(j),
\end{equation}
where $R(k)$, chosen such that $\max {U^k}^\T(j) R(k) U^k(j) << \max \rho^k(j)$, places a time-varying penalty on control actions which in turn ensures passenger comfort.

\begin{remark}
    $R(k)$ is allowed to be time varying so that its value is set inversely proportional to the sum of normalized instantaneous and predictive collision avoidance risks i.e.
    $$\frac{1}{R(k)} \propto \frac{2\kappa}{\kappa_a + \kappa_d} + \frac{2\tau}{\tau_a + \tau_d},$$
\end{remark}
which ensures that the emphasis on control action minimization (or passenger comfort) is further decreased with an increased risk of collision with other agents.

\subsection*{Complete Optimization Problem}
The receding horizon optimization problem is posed as a nonlinear program with its formulation provided below:
\begin{align}
    \min_{\mathcal{X}^k, \ \mathcal{U}^k}\;&J(k)\\
    \textbf{subject to:}&\nonumber\\
    X^k(0) &= X(k)\\
    X^k(j+1) &= f(X^k(j),U^k(j)) \quad \forall j \in \mathbb{Z}_{[0,H-1]},\\
    X^k(j) &\in \mathbb{X}(k) \quad \forall j \in \mathbb{Z}_{[0,H-1]}, \\
    U^k(j) &\in \mathbb{U}(k) \quad \forall j \in \mathbb{Z}_{[0,H-1]},
\end{align}
where $\mathcal{X}^k = \{X^k(j) \mid j \in \mathbb{Z}_{[0,H]} \}$ and $\mathcal{U}^k = \{U^k(j) \mid j \in \mathbb{Z}_{[0,H-1]} \}$, respectively representing the future states and controls, denote the optimization variables, while $f(X^k(j),U^k(j))$ serves as a compact representation of the system dynamics outlined in (\ref{eqn:dynamics}).

\begin{remark}
    A crucial requirement for timely collision avoidance and crash severity mitigation is a high algorithmic computational efficiency which is achieved by the smoothness and continuity of the objective function \cite{boydConvex}, as verified in Section \ref{sec:results}.
\end{remark}

\begin{remark}
    Our experimentation showed a minimal improvement in the computational efficiency of the non-linear program upon linearizing the dynamics. Therefore, we opt to use the non-linear dynamical model as its higher fidelity ensures the feasibility of outputs even in high-risk situations.
\end{remark}

\begin{figure*}[ht]
    \centering
    \includegraphics[width=0.3\textwidth]{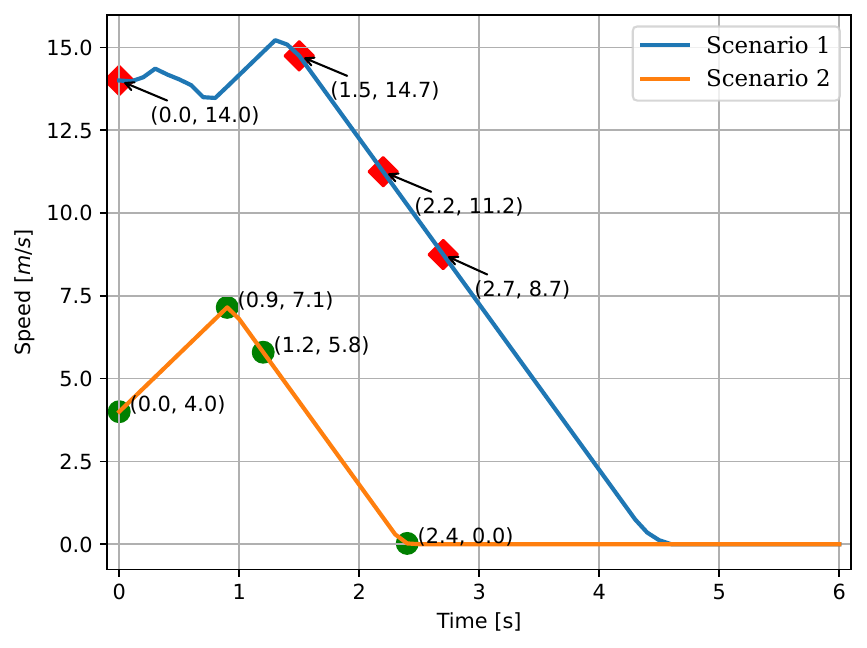}\label{fig:speed}
    \includegraphics[width=0.3\textwidth]{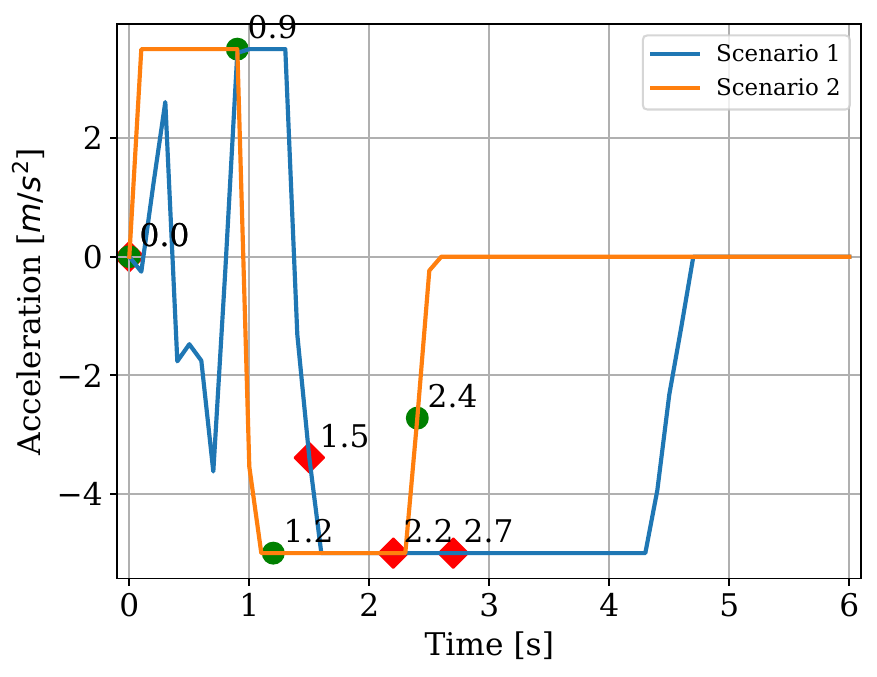}\label{fig:acceleration}
    \includegraphics[width=0.3\textwidth]{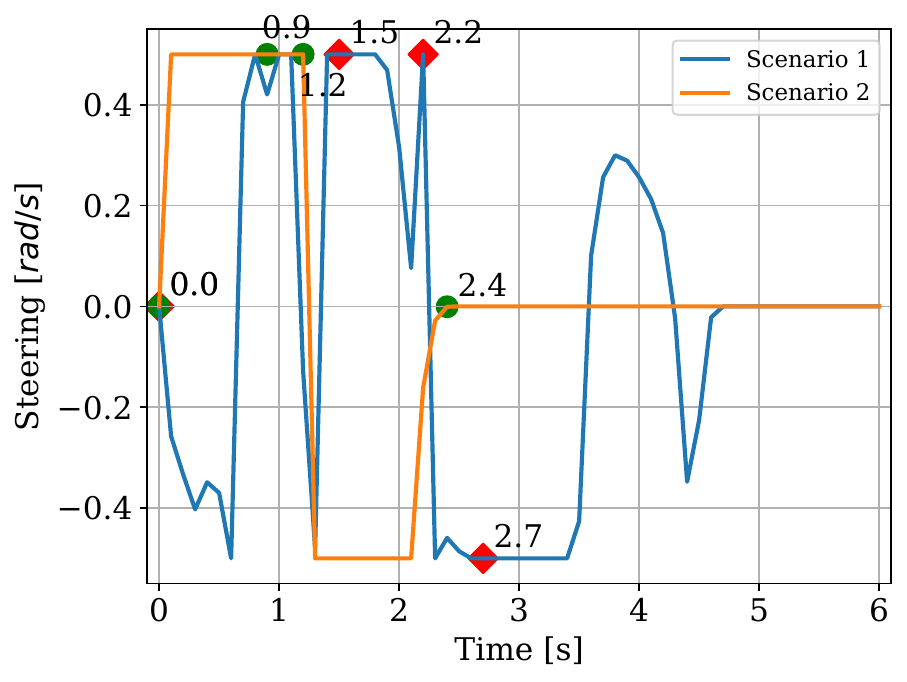}\label{fig:steering}
    
    \caption{\textbf{Control Plots for Scenarios 1 and 2.} The actuation limits are given by $U_{\text{min}} = \begin{bmatrix}
        -5 & -0.5
    \end{bmatrix}^\T$ and $U_{\text{max}} = \begin{bmatrix}
        3.5 & 0.5
    \end{bmatrix}^\T$, with negative steering ($\delta$) values representing steering to the right. In scenario 1, the ego vehicle initially switches between acceleration and deceleration while assessing the situation before deciding to slam on the brakes and steer aggressively to avoid the laterally moving object as well as the stopping vehicles. In scenario 2, the ego vehicle operates at the steering and acceleration limits to barely escape the rear-end speeding vehicle. The numbers indicated on the plots highlight the values at different time instances recorded in Fig. \ref{fig:scenario_animal} and Fig. \ref{fig:scenario_vehicle}.}
    \label{fig:scenarios_analysis}
\end{figure*}

\section{Evaluation}
This section details the experimental setup and demonstrates the performance of RCMS in our test case scenarios. 

\subsection{Experimental Setup} \label{sec:setup}
The experimentation is conducted on a computer equipped with an AMD Ryzen 7 5800h × 16 processor and NVIDIA GeForce RTX 3080 graphics card, running Ubuntu 20.04 LTS. To solve the non-linear program for RCMS, we employ the interior point optimizer \emph{IPOPT} \cite{ipopt} using \emph{MA27} linear solver, from the HSL library \cite{hsl}, within a CasADi \cite{casadi} environment. The optimization process requires $0.06s$ on average and $0.08s$ in the worst case, with $H=30$ and $T_s = 0.1$, showcasing the potential for real-time applicability.

\begin{figure} [ht]
\centering
\includegraphics[trim=0 0 0 0, clip,width=0.8\columnwidth]{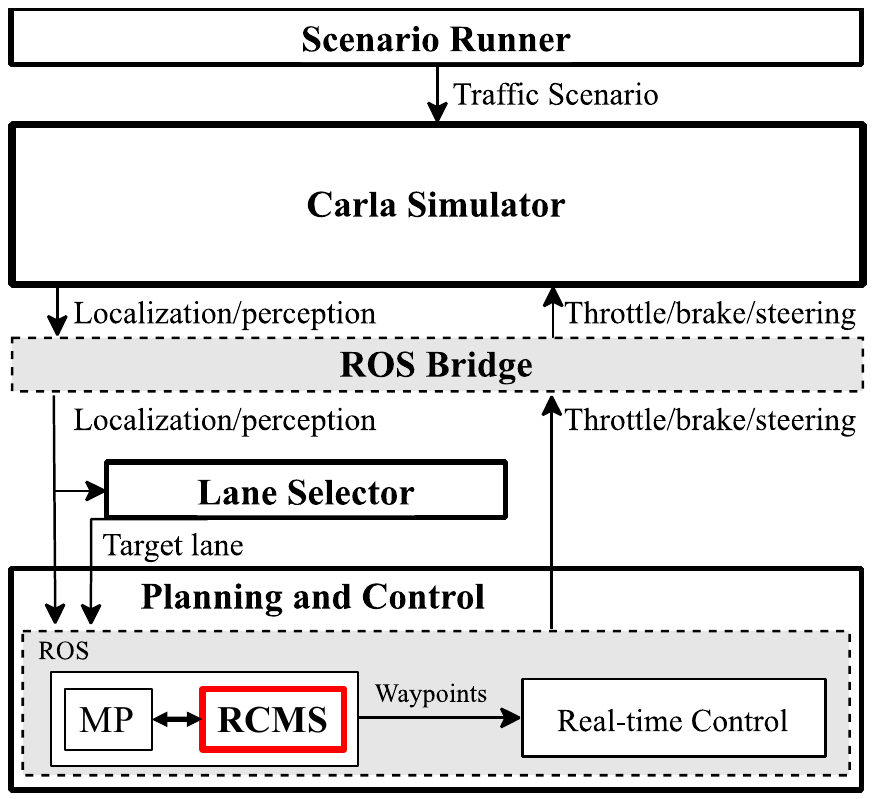}
\caption{\textbf{CARLA Simulation Setup}. Scenario Runner configures the scenario for the CARLA Simulator, which communicates with the Lane Selector as well as the Planning and Control ROS nodes via the ROS bridge node. The Planning and Control node incorporates both the MP and the RCMS modules, with the activation mechanism determining which of the two is operational.}
\label{fig:setup}
\vspace{-15pt}
\end{figure}

For the CARLA-based simulation in Section~\ref{sec:results_carla}, the experimental setup is depicted in Fig.~\ref{fig:setup}. It is composed of the CARLA Simulator and Scenario Runner (Versions 0.9.11) \cite{carla}, Lane Selector module \cite{mobil}, RCMS module (Section \ref{sec:approach}), and MP and Controller modules \cite{nnmpc}.


\subsection{Results} \label{sec:results}
We first demonstrate our approach on scenarios developed in a Python-based simulation environment in Section~\ref{sec:results_python} before moving on to the demonstration in a CARLA simulation environment in Section~\ref{sec:results_carla}.

\subsubsection{Python Simulation}   \label{sec:results_python}

To evaluate the performance of RCMS, we consider two scenarios: \begin{enumerate*}[label=(\roman*)] \item Scenario I: Object approaching laterally (Fig. \ref{fig:scenario_animal}); \item Scenario II: Fast-approaching rear-end vehicle (Fig. \ref{fig:scenario_vehicle})\end{enumerate*}. These scenarios are designed to evaluate different properties of RCMS. Scenario I tests whether the proposed framework has the ability to perform multiple evasive actions successively while Scenario II evaluates the ability of RCMS to operate at the actuation limits to minimize the severity of a potential collision. The corresponding speed, acceleration, and steering profiles are illustrated in Fig.~\ref{fig:scenarios_analysis}.

In Scenario I, with the object approaching laterally from the end of the road, the ego vehicle first assesses the situation to see if it can brake and swerve left to go behind the object. However, with the imposed actuation limits making that impossible, it then decides to accelerate and swerve right, while accessing the shoulder, to escape the object from its front at $t=1.5 s$. Upon noticing the successively stopping vehicles thereafter, it decides to brake and swerve left to avoid those vehicles before ending up in a safe zone at $t=2.7 s$ and handing the control back to MP. In Scenario II, the ego has to first accelerate while swerving left to avoid the fast-approaching rear-end vehicle. Once it barely escapes at around $t= 1s$, it slams on the brake while steering back right to avoid ramming into the road barrier before reaching a safe state at $t= 2.4 s$ and handing the control back to MP. 

\subsubsection{CARLA Simulation}  \label{sec:results_carla}

\begin{figure} [!htb]
\centering
\includegraphics[trim=0 0 0 0, clip,width=\columnwidth]{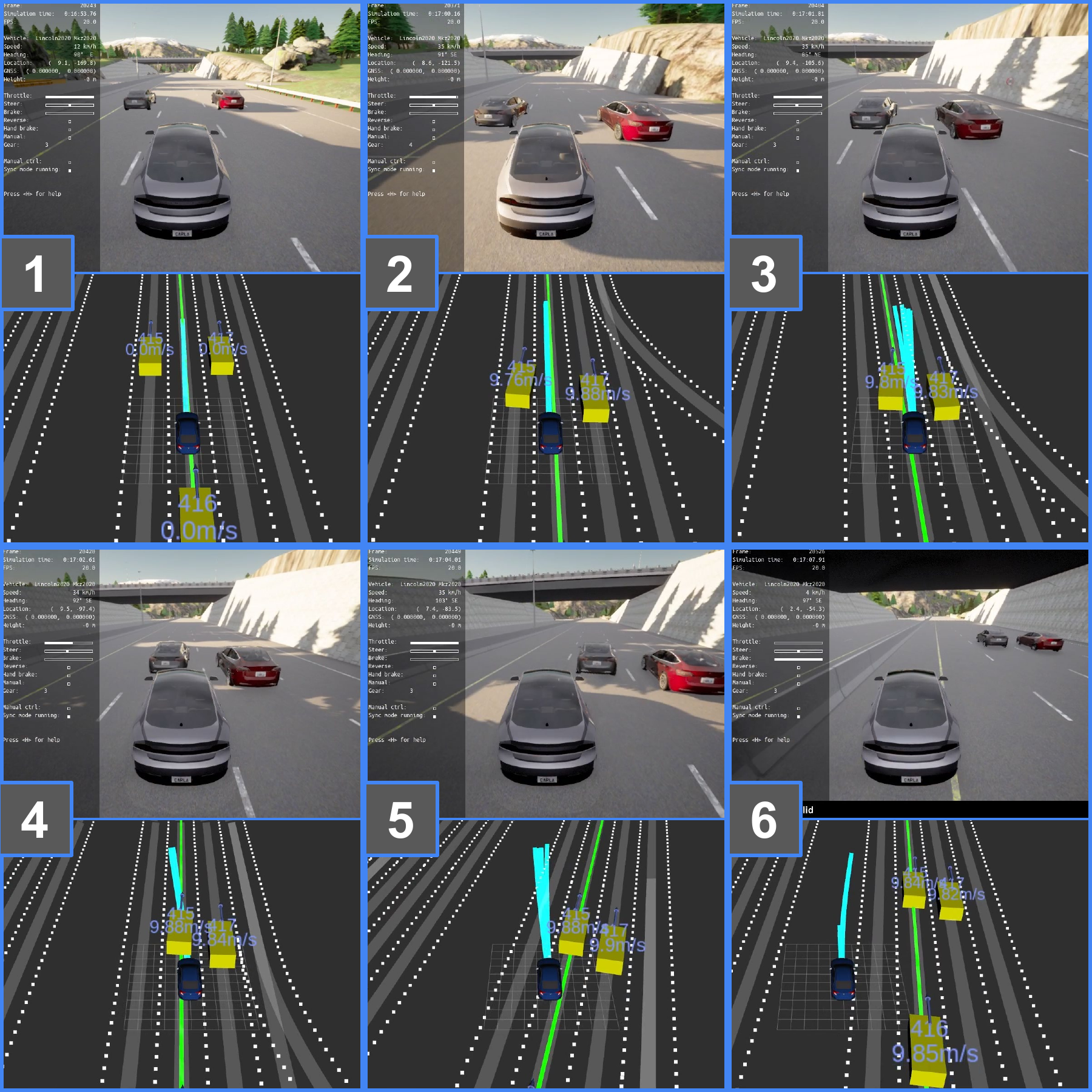}
\caption{\textbf{CARLA Simulation Scenario}. On a stretch of a multi-lane highway, lanes 0 (left), 1 (center), 2 (right), and 3 (right-most) are all available for traveling, while shoulder access is also available for emergency situations. The ego vehicle's motion over the course of the simulation is depicted through the numbered frames. The turquoise-colored lines represent the ego vehicle's planned trajectory at any given instant.}
\label{fig:scenarioCarla}
\end{figure}

To effectively evaluate the utility of RCMS in a realistic setting, we implement the motivational scenario outlined in Fig.~\ref{fig:overview} in a CARLA simulation environment. The scenario implementation, demonstrated in Fig.~\ref{fig:scenarioCarla}, consists of a four-lane highway segment, with lanes labeled as $0-3$ from left to right, relative to the direction of traffic. The ego vehicle, as depicted in Frame 1, is initialized to travel on lane $1$ with no leading vehicle, a tailgating vehicle, and two relatively slow vehicles traveling ahead on adjacent lanes $0$ and $2$. As the ego vehicle is approaching to overtake the vehicles in adjacent lanes, the vehicle in lane $0$ suddenly swerves in front of the ego vehicle, as shown in Frame 2, rendering the MP infeasible. In the absence of the RCMS module, the two backup strategies for MP, i.e. emergency braking and continuing on the current trajectory, both lead to a collision. In the case of emergency braking, there is a collision with the trailing vehicle while in the case of continuing on the current trajectory, there is a collision with the vehicle swerving in from lane $0$.

With the integration of RCMS, the ego vehicle carefully navigates around the surrounding vehicles, considering space availability and actuation limits, to safely avoid a collision, as depicted through Frames $3-6$. First, the ego vehicle cautiously maneuvers to the right to see if it can exploit the available space to pass between the vehicles, as observed in Frame 3. However, with the actuation limits preventing that, it decides to slow down slightly to accommodate the swerving vehicle, as seen in Frame 4. Wary of the negligible distance to the now leading vehicle, RCMS carefully steers the ego vehicle towards the left, as seen in Frame 5, to place it in a safe state while accessing the shoulder, as seen in Frame 6, before handing the control back to MP. This highlights the efficacy of RCMS in handling challenging collision-prone scenarios.

\section{Conclusion}
In conclusion, we have developed a novel risk-aware crash mitigation system that comprises an activation mechanism and a modular trajectory generation component to perform evasive maneuvers in high-risk collision-prone situations. The activation mechanism effectively combines instantaneous and predictive collision risk evaluation within a hysteresis band to facilitate a smooth transition between RCMS and MP, with their distinct objectives, to maintain passenger comfort. Meanwhile, the trajectory generation component minimizes situational risk through a smooth function while considering actuation, dynamical, and road limits. We have validated the real-time applicability and performance of our approach by conducting simulations of two high-risk scenarios which assessed the ability to perform successive evasive maneuvers at the vehicle's actuation limits. 
Our future work entails the implementation of RCMS on a physical small-scale setup to evaluate its robustness to various real-world uncertainties before deploying it on actual vehicles.

\bibliographystyle{IEEEtran}
\bibliography{references}

\begin{thebibliography}{10}
\providecommand{\url}[1]{#1}
\csname url@samestyle\endcsname
\providecommand{\newblock}{\relax}
\providecommand{\bibinfo}[2]{#2}
\providecommand{\BIBentrySTDinterwordspacing}{\spaceskip=0pt\relax}
\providecommand{\BIBentryALTinterwordstretchfactor}{4}
\providecommand{\BIBentryALTinterwordspacing}{\spaceskip=\fontdimen2\font plus
\BIBentryALTinterwordstretchfactor\fontdimen3\font minus
  \fontdimen4\font\relax}
\providecommand{\BIBforeignlanguage}[2]{{%
\expandafter\ifx\csname l@#1\endcsname\relax
\typeout{** WARNING: IEEEtran.bst: No hyphenation pattern has been}%
\typeout{** loaded for the language `#1'. Using the pattern for}%
\typeout{** the default language instead.}%
\else
\language=\csname l@#1\endcsname
\fi
#2}}
\providecommand{\BIBdecl}{\relax}
\BIBdecl

\bibitem{emergencyBraking}
I.~Isaksson-Hellman and M.~Lindman, ``Evaluation of rear-end collision
  avoidance technologies based on real world crash data,'' \emph{Proceedings of
  the Future Active Safety Technology Towards zero traffic accidents
  (FASTzero), Gothenburg, Sweden}, pp. 9--11, 2015.

\bibitem{fuzzyRear}
V.~Milan{\'e}s, J.~P{\'e}rez, J.~Godoy, and E.~Onieva, ``A fuzzy aid rear-end
  collision warning/avoidance system,'' \emph{Expert Systems with
  Applications}, vol.~39, no.~10, pp. 9097--9107, 2012.

\bibitem{huijser2009animal}
M.~P. Huijser, T.~D. Holland, A.~V. Kociolek, A.~M. Barkdoll, J.~D. Schwalm
  \emph{et~al.}, ``Animal-vehicle crash mitigation using advanced technology:
  phase ii, system effectiveness and system acceptance.'' Oregon. Dept. of
  Transportation. Research Unit, Tech. Rep., 2009.

\bibitem{muller2018machine}
M.~M{\"u}ller, M.~Botsch, D.~B{\"o}hml{\"a}nder, and W.~Utschick, ``Machine
  learning based prediction of crash severity distributions for mitigation
  strategies,'' \emph{Journal of Advances in Information Technology}, vol.~9,
  no.~1, pp. 15--24, 2018.

\bibitem{samplingPOM}
K.~Lee and D.~Kum, ``Collision avoidance/mitigation system: Motion planning of
  autonomous vehicle via predictive occupancy map,'' \emph{IEEE Access},
  vol.~7, pp. 52\,846--52\,857, 2019.

\bibitem{wang2019crash}
H.~Wang, Y.~Huang, A.~Khajepour, Y.~Zhang, Y.~Rasekhipour, and D.~Cao, ``Crash
  mitigation in motion planning for autonomous vehicles,'' \emph{IEEE
  transactions on intelligent transportation systems}, vol.~20, no.~9, pp.
  3313--3323, 2019.

\bibitem{shang2023emergency}
X.~Shang and A.~Eskandarian, ``Emergency collision avoidance and mitigation
  using model predictive control and artificial potential function,''
  \emph{IEEE Transactions on Intelligent Vehicles}, 2023.

\bibitem{integratedCrashAvoidanceMitigation}
Y.~Qin, E.~Hashemi, and A.~Khajepour, ``Integrated crash avoidance and
  mitigation algorithm for autonomous vehicles,'' \emph{IEEE Transactions on
  Industrial Informatics}, vol.~17, no.~11, pp. 7246--7255, 2021.

\bibitem{surveyPlanning}
B.~Paden, M.~{\v{C}}{\'a}p, S.~Z. Yong, D.~Yershov, and E.~Frazzoli, ``A survey
  of motion planning and control techniques for self-driving urban vehicles,''
  \emph{IEEE Transactions on intelligent vehicles}, vol.~1, no.~1, pp. 33--55,
  2016.

\bibitem{slas}
F.~M. Tariq, D.~Isele, J.~S. Baras, and S.~Bae, ``Slas: Speed and lane advisory
  system for highway navigation,'' in \emph{2022 61st IEEE Conference on
  Decision and Control (CDC)}, 2022.

\bibitem{overtakingBidirectional}
F.~M. Tariq, N.~Suriyarachchi, C.~Mavridis, and J.~S. Baras, ``Autonomous
  vehicle overtaking in a bidirectional mixed-traffic setting,'' in \emph{2022
  American Control Conference (ACC)}.\hskip 1em plus 0.5em minus 0.4em\relax
  IEEE, 2022, pp. 3132--3139.

\bibitem{cooperativeOvertaking}
------, ``Cooperative bidirectional mixed-traffic overtaking,'' in \emph{2022
  IEEE 25th International Conference on Intelligent Transportation Systems
  (ITSC)}.\hskip 1em plus 0.5em minus 0.4em\relax IEEE, 2022, pp. 2494--2501.

\bibitem{bicycleModel}
J.~Kong, M.~Pfeiffer, G.~Schildbach, and F.~Borrelli, ``Kinematic and dynamic
  vehicle models for autonomous driving control design,'' in \emph{2015 IEEE
  intelligent vehicles symposium (IV)}.\hskip 1em plus 0.5em minus 0.4em\relax
  IEEE, 2015, pp. 1094--1099.

\bibitem{tireModel}
Y.~Gao, T.~Lin, F.~Borrelli, E.~Tseng, and D.~Hrovat, ``Predictive control of
  autonomous ground vehicles with obstacle avoidance on slippery roads,'' in
  \emph{Dynamic systems and control conference}, vol. 44175, 2010, pp.
  265--272.

\bibitem{riskGaussian}
A.~Moradipari, S.~Bae, M.~Alizadeh, E.~M. Pari, and D.~Isele, ``Predicting
  parameters for modeling traffic participants,'' in \emph{2022 IEEE 25th
  International Conference on Intelligent Transportation Systems (ITSC)}.\hskip
  1em plus 0.5em minus 0.4em\relax IEEE, 2022, pp. 703--708.

\bibitem{threatAssessment}
Y.~Zhang, E.~K. Antonsson, and K.~Grote, ``A new threat assessment measure for
  collision avoidance systems,'' in \emph{2006 IEEE Intelligent Transportation
  Systems Conference}.\hskip 1em plus 0.5em minus 0.4em\relax IEEE, 2006, pp.
  968--975.

\bibitem{mpcDriving}
P.~Falcone, F.~Borrelli, J.~Asgari, H.~E. Tseng, and D.~Hrovat, ``Predictive
  active steering control for autonomous vehicle systems,'' \emph{IEEE
  Transactions on Control Systems Technology}, vol.~15, no.~3, pp. 566--580,
  2007.

\bibitem{skewNormal}
A.~Azzalini and A.~D. Valle, ``The multivariate skew-normal distribution,''
  \emph{Biometrika}, vol.~83, no.~4, pp. 715--726, 1996.

\bibitem{surveyPrediction}
S.~Lef{\`e}vre, D.~Vasquez, and C.~Laugier, ``A survey on motion prediction and
  risk assessment for intelligent vehicles,'' \emph{ROBOMECH journal}, vol.~1,
  no.~1, pp. 1--14, 2014.

\bibitem{humanError}
D.~Sam, C.~Velanganni, and T.~E. Evangelin, ``A vehicle control system using a
  time synchronized hybrid vanet to reduce road accidents caused by human
  error,'' \emph{Vehicular comm.}, vol.~6, pp. 17--28, 2016.

\bibitem{crashStats}
S.~Singh, ``Critical reasons for crashes investigated in the national motor
  vehicle crash causation survey,'' Tech. Rep., 2015.

\bibitem{boydConvex}
S.~Boyd, S.~P. Boyd, and L.~Vandenberghe, \emph{Convex optimization}.\hskip 1em
  plus 0.5em minus 0.4em\relax Cambridge university press, 2004.

\bibitem{ipopt}
A.~W{\"a}chter and L.~T. Biegler, ``On the implementation of an interior-point
  filter line-search algorithm for large-scale nonlinear programming,''
  \emph{Mathematical programming}, vol. 106, pp. 25--57, 2006.

\bibitem{hsl}
I.~S. Duff, ``Sparse system solution and the hsl library,'' \emph{Some topics
  in industrial and applied mathematics}, vol.~8, pp. 78--94, 2006.

\bibitem{casadi}
J.~A.~E. Andersson, J.~Gillis, G.~Horn, J.~B. Rawlings, and M.~Diehl,
  ``{CasADi} -- {A} software framework for nonlinear optimization and optimal
  control,'' \emph{Mathematical Programming Computation}, vol.~11, no.~1, pp.
  1--36, 2019.

\bibitem{carla}
A.~Dosovitskiy, G.~Ros, F.~Codevilla, A.~Lopez, and V.~Koltun, ``Carla: An open
  urban driving simulator,'' in \emph{Conference on robot learning}.\hskip 1em
  plus 0.5em minus 0.4em\relax PMLR, 2017, pp. 1--16.

\bibitem{mobil}
\BIBentryALTinterwordspacing
A.~Kesting, M.~Treiber, and D.~Helbing, ``General lane-changing model mobil for
  car-following models,'' \emph{Transportation Research Record}, vol. 1999,
  no.~1, pp. 86--94, 2007. [Online]. Available:
  \url{https://doi.org/10.3141/1999-10}
\BIBentrySTDinterwordspacing

\bibitem{nnmpc}
S.~Bae, D.~Saxena, A.~Nakhaei, C.~Choi, K.~Fujimura, and S.~Moura,
  ``Cooperation-aware lane change maneuver in dense traffic based on model
  predictive control with recurrent neural network,'' in \emph{2020 American
  Control Conference (ACC)}.\hskip 1em plus 0.5em minus 0.4em\relax IEEE, 2020,
  pp. 1209--1216.

\end{thebibliography}

\end{document}